\documentclass[letterpaper, 10 pt, conference]{ieeeconf}  % Comment this line out if you need a4paper

\IEEEoverridecommandlockouts                
\usepackage{graphicx}
\usepackage{float}

\usepackage{todonotes}
\usepackage{booktabs}
\renewcommand{\baselinestretch}{1}       % for squeezing the draft into the page limit, do not use

\def\etal{et~al.~}			  % and others, and co-workers
\def\eg{e.g.~}               % for example
\def\ie{i.e.,~}               % that is, in other words
\newlength\paramargin
\newlength\figmargin
\newlength\secmargin

\newlength\figwidth

\newcommand{\figref}[1]{Figure~\ref{fig:#1}}

\long\def\ignorethis#1{}

\usepackage{amsmath}
\usepackage{hyperref}

\makeatletter
\let\NAT@parse\undefined
\makeatother
\usepackage[numbers]{natbib}
\title{\LARGE \bf Learning to See before Learning to Act: \\ Visual Pre-training for Manipulation}
\author{Lin Yen-Chen${^{12}}$ Andy Zeng${^1}$ Shuran Song${^{13}}$ Phillip Isola${^2}$ Tsung-Yi Lin${^1}$% <-this % stops a space
\vspace{0.1cm} \\
$^{1}$Google, Brain team $^{2}$Massachusetts Institute of Technology $^{3}$Columbia University%
}

\begin{document}

\maketitle
\thispagestyle{empty}
\pagestyle{empty}

%%%%%%%%%%%%%%%%%%%%%%%%%%%%%%%%%%%%%%%%%%%%%%%%%%%%%%%%%%%%%%%%%%%%%%%%%%%%%%%%
\begin{abstract}
Does having visual priors (\eg the ability to detect objects) facilitate learning to perform vision-based manipulation (\eg picking up objects)? We study this problem under the framework of transfer learning, where the model is first trained on a passive vision task (\ie the data distribution does not depend on the agent's decisions), then adapted to perform an active manipulation task (\ie the data distribution \emph{does} depend on the agent's decisions).
We find that pre-training on vision tasks significantly improves generalization and sample efficiency for learning to manipulate objects. However, realizing these gains requires careful selection of which parts of the model to transfer. Our key insight is that outputs of standard vision models highly correlate with \emph{affordance maps} commonly used in manipulation. Therefore, we explore directly transferring model parameters from vision networks to affordance prediction networks, and show that this can result in successful zero-shot adaptation, where a robot can pick up certain objects with zero robotic experience. With just a small amount of robotic experience, we can further fine-tune the affordance model to achieve better results. With just 10 minutes of suction experience or 1 hour of grasping experience, our method achieves $\sim 80\%$ success rate at picking up novel objects. Experiment videos can be found at \url{http://yenchenlin.me/vision2action/}.

\end{abstract}

%%%%%%%%%%%%%%%%%%%%%%%%%%%%%%%%%%%%%%%%%%%%%%%%%%%%%%%%%%%%%%%%%%%%%%%%%%%%%%%%
\section{Introduction}

The development of embodied cognition happens incrementally~\cite{smith2005development}, with infants' early experiences strongly ordered by the stages of development of their sensorimotor systems. After 3 or 4 months of looking at the world \textit{passively}, infants begin to reach for objects \textit{actively}. This staged nature of development organizes experiences in a natural curriculum that shifts supervisory signals from observation to interaction.% -- illustrating the presence of a curriculum that gradually introduces more complex concepts.
%

%
%Inspired by this, classical solutions for robot picking typically include learning the perception module from offline observations. However, state-of-the-art perception models which employ deep neural networks often require tens to hundreds of thousands of labeled training examples to achieve adequate performance. The hurdle of data labelling.
%

%
In contrast, recent breakthroughs on vision-based manipulation~\cite{pinto2016supersizing,zeng2019tossingbot,song2019grasping,zeng2019learning} often ignore the staged nature of cognitive development and instead directly supervise control policies via interactions between robots and environments.
This causes two major problems. First, the learning process is slow and expensive: learning a new manipulation skill usually demands hundreds to thousands of real-world interactions. Second, the learned policy does not generalize well to unseen objects, because it has only interacted with a limited number of training objects in a specific environment. 

\begin{figure}[t]
    \centering
    \includegraphics[width=\linewidth]{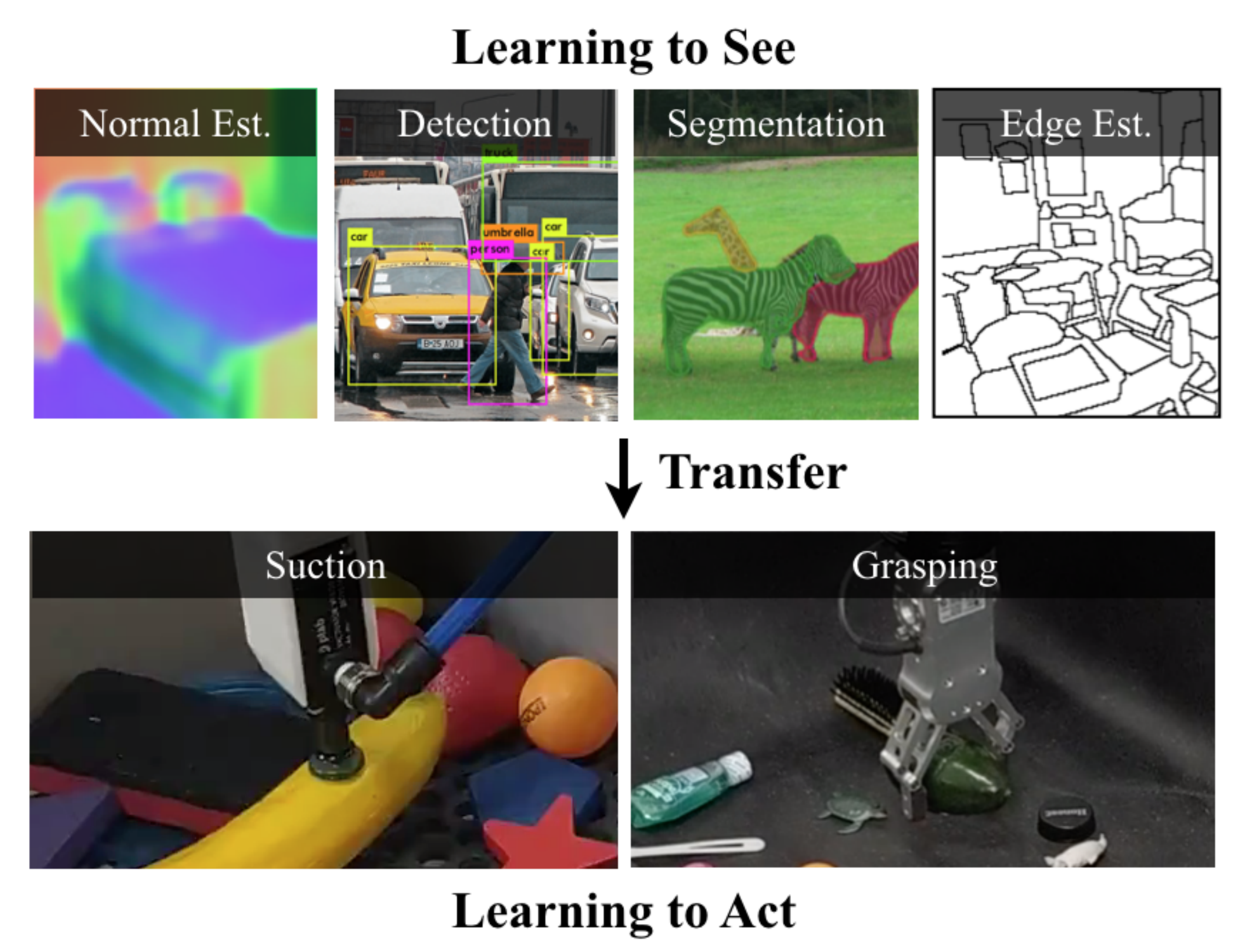}
     \vspace{-8mm}
      %of learning to see before learning to act
    \caption{ {\bf Overview.} Does first learning to see aid the speed at which a robot can learn to act? In this paper, we show how to transfer knowledge from pre-trained vision tasks to robotic manipulation policies, demonstrating large improvements in the sample efficiency and generalization ability of robot learning.
    %{\bf Overview:} We study if/how a model pre-trained on passive vision tasks can transfer to robotic manipulation tasks. 
    %No pre-training corresponds to directly learning affordance maps from robot-environment interactions~\cite{zeng2018learning,zeng2019tossingbot}. 
    %We report significant improvement in sample efficiency and generalization when pre-training on passive vision tasks.
    %
    }
    \label{fig:overview}
    \vspace{-6mm}
\end{figure}

In this paper, we aim to answer the question: does having visual priors (\eg the ability to detect objects) facilitate learning to perform vision-based manipulation (\eg picking up objects)?
We adopt the framework of transfer learning, where the model is first trained on a source task, and then reused to fit the target task. Specifically, we study the problem of transferring a vision model learned from visual observations to an affordance model for a manipulation task. Although transfer learning has already shown many promising results on vision tasks, it remains unclear how to transfer knowledge learned from \textit{passive} vision tasks (\ie the data distribution is independent of the agent's decisions)  to \textit{active} robotics tasks (\ie the data distribution depends on the agent's decisions), where good exploration strategies are vital.
%

% The first puzzling question to ask for transferring vision models to perform robotics tasks is: Which parts of the model should we transfer? In transfer learning, the most common approach is to transfer learned model parameters. This assumes a common model architecture shared for both source and target models. Typically, a backbone (e.g., ResNet) is shared and each task has a few task specific convolution or linear layers on top of the backbone, coined prediction head. Therefore, most transfer learning work focuses on transferring latent feature representations from backbone and let the prediction head be randomly initialized. Zeng et al.~\cite{zeng2018learning} shows that transferring latent feature representations from model pre-trained on ImageNet does not accelerate learning of action policy or converge to better performance.
%
One obvious solution to this problem is to transfer the parameters of a shared model architecture, usually termed a ``backbone'', while randomly initializing task-specific convolutional or linear layers, which are commonly called ``heads'', on top of it. Essentially, this approach transfers the pre-trained latent feature representations of vision models to affordance models. However, Zeng et al.~\cite{zeng2018learning} found that transferring latent feature representations from a model pre-trained on ImageNet does not accelerate the learning process or converge to better performance for pushing-grasping tasks.%synergies.

In our work, we find that only transferring latent features for learning affordance models results in bad \textit{exploration}. The key problem is that by randomly initializing the head of the affordance model, the resulting policy, even with pre-trained latent features, still \textit{randomly} explores the environments and thus fails to collect useful supervisory signals. On the contrary, humans are able to take advantage of a set of perceptual biases (\eg object-centric world structure) to aid the exploration of possible interactions in the environments. For example, babies do not randomly try all possible actions, grasping haphazardly at the air; they are biased toward interacting preferentially with the most obvious objects~\cite{johnson2003development}. Inspired by this observation, we propose to transfer the entire vision model, including both features from the backbone and the visual predictions from the head, to initialize the affordance model. In this case, the resulting initial policy is simply the visual prediction model pre-trained on passive observations. 

We show that such vision-guided exploration greatly reduces the number of interactions needed to acquire a new manipulation skill,  then we explore which visual predictions can be used as a good initialization for the affordance model. We experiment with a wide variety of vision tasks and find that initializing the affordance model with most of them results in significantly faster convergence speed compared to a randomly initialized model, which suggests that many vision tasks can be helpful for learning the affordance model. 
%In practice, however, training a vision model for each target environment would be very costly due to the efforts at data collection. To address this problem, we propose a simple strategy to turn an off-the-shelf vision model into the initialization of an affordance model without the need for data collection in the target domain. 
%In real-world robotic experiments, we show that our proposed method greatly enhances the affordance model's sample efficiency and generalization. 
Based on our results, we believe that learning to see from abundant visual observations before learning to act from expensive robot-environment interactions is a key step to improve the generalization ability and sample efficiency of vision-based manipulation.

%
%In our work, we first show that instead of following the common practice in computer vision, which only transfers the features and randomly initialized the prediction head, transferring both features and prediction head dramatically improve the sample efficiency  due to better exploration provided by vision models' predictions. Second, we benchmark the affinity between different vision and manipulation tasks to better understand the underlying structure among these tasks.
%Then, we perform real world experiments to verify our findings and propose a simple strategy to turn an off-the-shelf vision model which encodes the object-centric prior into the initialization for affordance prediction. We show that our initialization method greatly enhances model's sample efficiency and generalization. 
%

%
Our contributions are three-fold: 1) we explore the problem of transferring vision models to perform robotics tasks and show that transferring both the backbone and the head of networks leads to better performance compared to only transferring the backbone, 2) we systematically benchmark the effectiveness of pre-training on different vision tasks for learning grasping and suction, 3) we offer a simple and powerful initialization strategy to turn an off-the-shelf vision model into the initialization of an affordance model without additional data collection. Specifically, this strategy enables a robot to learn to perform suction in 10 minutes and grasping in an hour with an approximately $80\%$ success rate on unseen objects.

\begin{figure*}[t]
    \centering
    \includegraphics[width=\linewidth]{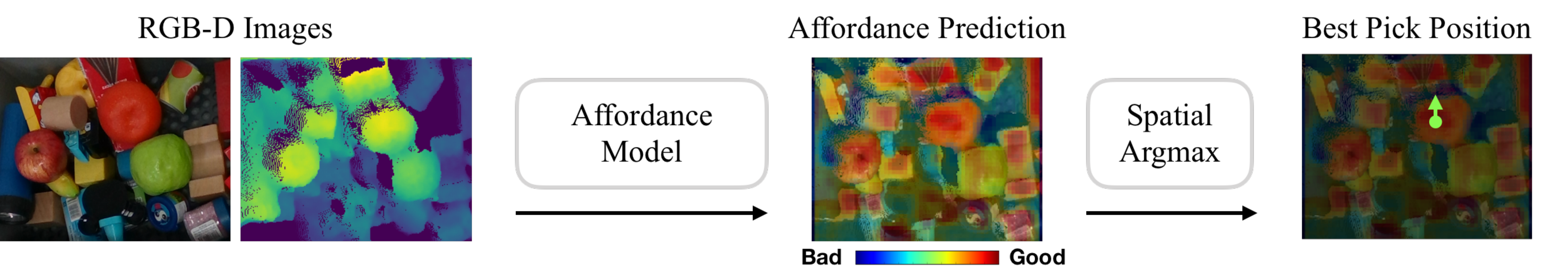}
    \caption{{\bf Illustration of affordance-based manipulation.} Given an RGB-D image, we predict pixel-wise affordance with an affordance model. At each pixel, the predicted value of affordance represents the success rate of performing the corresponding motion primitive at that pixel's projected 3D position. The motion primitive is executed by the robot at the position with the highest affordance value.
    }
    \label{fig:affordance}
\end{figure*}
\section{Related Work}

% \begin{figure*}[ht]
%     \centering
%     \includegraphics[width=\linewidth]{figures/affordance.png}
%     \caption{{\bf Place holder for explaining affordance.} 
%     %
%     }
%     \label{fig:affordance}
% \end{figure*}

\noindent \textbf{Offline Computer Vision.}
Computer vision tasks focus on estimating environment properties, e.g. image classification~\cite{krizhevsky2012imagenet}, detection~\cite{lin2017focal}, segmentation~\cite{he2017mask,song2018im2pano3d}, and pose estimation~\cite{cao2017realtime}. 
Although they focus on estimating different latent properties, the learning process happens on prerecorded datasets. 
In this work, we study how to transfer the knowledge learned from prerecorded vision datasets to solve downstream manipulation tasks that require agents to collect their own datasets actively.

\noindent \textbf{Self-supervised Robot Learning.}
Over the last few years, there has been growing interest in scaling up robot data collection through self-supervised learning.
This approach has been applied to learn different robotic skills including grasping~\cite{pinto2016supersizing,pinto2016curious,levine2018learning,calandra2017feeling,zakka2019form2fit}, pushing~\cite{finn2017deep,agrawal2016learning,ebert2017self}, tossing~\cite{zeng2019tossingbot}, and obstacle avoidance~\cite{kahn2017uncertainty,gandhi2017learning}.
Specifically for grasping, Pinto~\etal~\cite{pinto2016supersizing} uses background subtraction to reduce the number of random grasping trials in empty space and shows that it improve sample efficiency.
Building on this prior work, we instead explore how transferring model weights can improve not only sample efficiency but also generalization to unseen objects.

\noindent \textbf{Transfer Learning.}
Previous works on transfer learning for robotics focus on reusing skills learned from different environments~\cite{rusu2016progressive} or bridging the gap between simulation and reality~\cite{tobin2017domain,tan2018sim,peng2018sim,rusu2016sim}.
In our work, we study the problem of how to transfer models learned from passive vision tasks to learn active manipulation tasks.

\noindent \textbf{Representation Learning.}
The field of unsupervised learning has explored different ways to learn state representations~\cite{higgins2017darla,ghosh2018learning,xu2019densephysnet} for policy learning.
Recently, several works~\citep{zhou2019does,sax2018mid,situational2019} have studied the benefits of combining various mid-level visual representations for reinforcement learning. 
Different from previous works, we 1) demonstrate real-world results on manipulation while previous works present simulation results on navigation, 2) highlight the importance of using a pre-trained model for exploration, and 3) propose an initialization strategy without the need of data collection at target domain.
\begin{figure}[t]
    \centering
    \includegraphics[width=\linewidth]{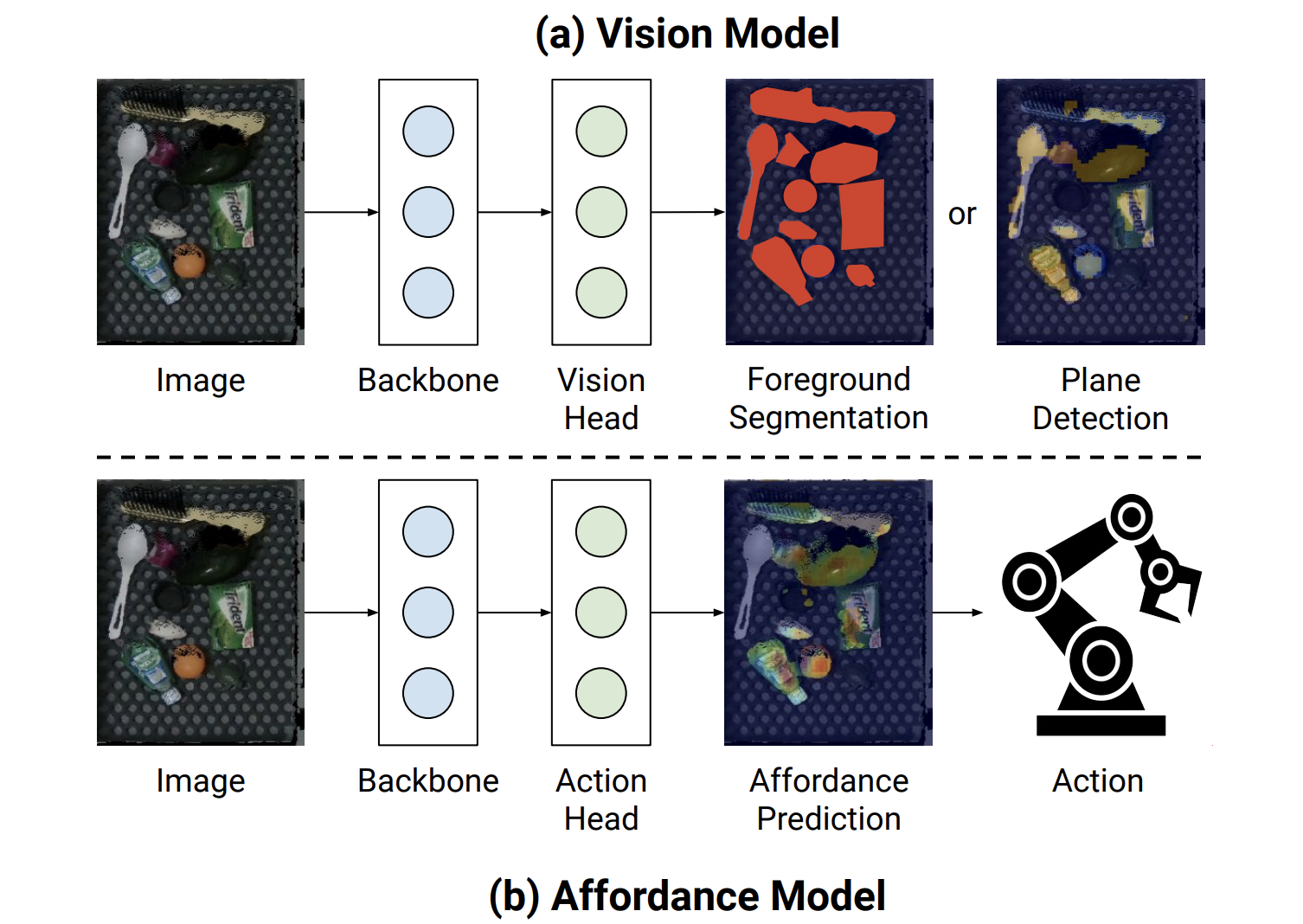}
    \caption{ {\bf Comparison of vision and affordance models.} Both vision and affordance models take RGB-D image as input and process it through backbone and head to output predictions. The major difference is that affordance models' predictions will then be executed by the robot with corresponding motion primitive to affect the future data distribution.
    } 
    \label{fig:model}
    \vspace{-4mm}
\end{figure}

\section{Method}
%%%
% TY: I think the method section is about how to do vision to action transfer. Now the section reads more like how to build a model to perform picking.
%%%
In this section, we introduce the input representation, the vision models, and manipulation affordance models (Figure~\ref{fig:model}), followed by the details of transfer learning.

\subsection{Input Representation}
We represent the visual input $I$ as an RGB-D heightmap image of the workspace (\ie a bin of objects). To obtain the heightmap, we follow the pre-processing steps in Zeng \etal~\cite{zeng2018learning} by first projecting the RGB-D image captured by the camera onto a 3D point cloud, and orthographically back-project upwards in the gravity direction. The RGB and D channels are both normalized (mean-subtracted and divided by standard deviation) before being fed into the model.
In our real-world setup for suction, images captured by the camera cover a $0.36 \times 0.26$m area and have a pixel resolution of $288 \times 208$. For grasping, we have the camera cover a $0.5 \times 0.5$m area with pixel resolution of $400 \times 400$. In both cases, each pixel represents a $1.25 \times 1.25$mm vertical column in the robot's workspace.

\subsection{Learning Vision Models}
We follow the common practice in~\cite{he2017mask} to design the vision model using ConvNet. The model has a heavy backbone model for extracting visual features and a lightweight vision head, which usually consists of one or a few convolution layers for making visual prediction from features of a backbone model. The visual prediction is a 2D heatmap that can represent a wide variety of vision tasks.

In this work, we consider the following vision tasks: 1) edge detection, 2) corner detection, 3) object center detection, 4) foreground segmentation, and 5) surface normal estimation. Specifically, we transform surface normal estimation as identifying pixels whose normal are similar to the anti-gravity direction. We collect the dataset by letting robots randomly interact with the environment. The annotations for each task are obtained through simple computer vision techniques such as Canny edge detection~\cite{canny1986}, Harris corner detection~\cite{harris88acombined}, and background subtraction. In our real-world setup, we also experiment with additional vision models trained with ImageNet classification and COCO object detection, which do not contain any domain and task-specific training data from the environment where the robot is deployed.

To train the vision system, we apply binary cross-entropy loss: $L = -y_i \log{f(I_i)} - (1-y_i) \log{(1-f(I_i))}$ ,where $I_i$ is the input RGB-D image, $y_i$ is the binary ground truth label, and $f$ is the vision model. Gradients are computed over all the pixels in the input RGB-D image.

% \begin{align}\label{binary_cross_entropy}
% L = - (y_i \log{f(I_i)} + (1-y_i) \log{1-f(I_i)})
% \end{align}

%

\subsection{Learning Manipulation Affordances}
The affordance model, which consists of a ConvNet and a motion primitive, is illustrated in~\figref{affordance}. The ConvNet learns to predict a dense 2D heatmap that encodes the probability of picking success at each pixel location. The motion primitive is a pre-defined function which controls the robot arm to perform manipulation task from a fixed initial arm position. Different manipulation tasks need their own motion primitive. The motion primitive is open-loop, with motion planning executed using a stable, collision-free IK solver.

\noindent \textbf{Suction primitive.} The suction primitive takes as input parameters $\phi_s = (p)$ and executes a top-down suction centered at a 3D position $p = (p_x, p_y, p_z)$. During execution, the suction cup approaches $p$ along the gravity direction until the 3D position of the suction cup meets $p$, at which point the suction activates, and lifts upwards 10cm. Picking success is then determined by thresholding on the difference between digital vacuum pressure readings before and after the primitive, which serves as a sufficient signal to determine whether the suction contact resulted in a good seal.

\noindent \textbf{Grasping primitive.} The grasping primitive takes as input parameters $\phi_g = (p, \theta)$ and executes a top-down parallel-jaw grasp centered at a 3D position $p = (p_x, p_y, p_z)$ oriented $\theta$ around the gravity direction. In order to get $\theta$, we rotate the input heightmap by 16 different angles and feed all of them into our model to estimate the affordance of horizontal grasps for each heightmap. We then perform a max-pooling across all heightmaps to identify the best grasping position and orientation. During execution, the open gripper approaches $p$ along the gravity direction until the 3D position of the middle point between the gripper fingertips meets $p$, then the gripper closes and lifts upwards 10cm. Picking success is determined by thresholding on the measured width between gripper fingertips. 

As the data for learning affordances comes from robot-environment interactions, it is not independent and identically distributed. Thus, we curate incoming data in an experience replay buffer and perform prioritized experience replay to construct inputs for each iteration of training. To train the affordance model, we again adopt binary cross-entropy loss. In this case, $I_i$ is the input RGB-D image, $y_i$ is the binary ground truth picking success label, and $f$ is the affordance model. Since we only have the label of the single pixel on which the picking primitive was executed, gradients are only passed through that pixel. All other pixels have zero loss.

\subsection{Transferring Vision Model to Affordance Model}
To facilitate the transfer, we design vision and affordance models to share the same backbone and head architecture. The learned parameters of the vision model are directly taken to initialize parameters of affordance model. We study two types of vision to affordance transfer: 1) transfer of backbone model; 2) transfer of backbone and head models. It is common to only transfer backbone model, especially when the head architectures are not compatible, e.g., image classification does not have the same architecture as affordance prediction head so one can only transfer the backbone of ImageNet pre-trained model. However, we show that transferring the entire vision model, including both backbone and head, are critical for improving sample efficiency for learning the affordance model. The intuition of transferring the head is that visual predictions are closely related to affordance of manipulation (see ~\figref{visualization}). For example, picking the object center can be a good manipulation policy for apple. In our experiments, we find that exploring and fine-tuning with respect to visual predictions greatly speed up the learning.

\section{Experiments}
\label{sec:experiments}
% \todo[inline]{I'd really like to have a figure shows heat map of heat map of random / different vision predictions / grasping and suction.}
% \todo[inline]{The points we'd like to answer in sim. experiments are: (1) show that transfer both backbone + head works better than just transfer the latent features in backbone; (2) what visual tasks transfer well to action tasks}
%\todo[inline]{What we want to show in real world experiments are: (1) the best vision model from sim also works well in real. So the points in sim are also valid in real; (2) passive observations from a different source domain (COCO) can help generalization in action tasks. This isn't shown in sim experiments; (3) comparison to the traditional robotics practice.}

\begin{figure}[t]
    \centering
    \includegraphics[width=\linewidth]{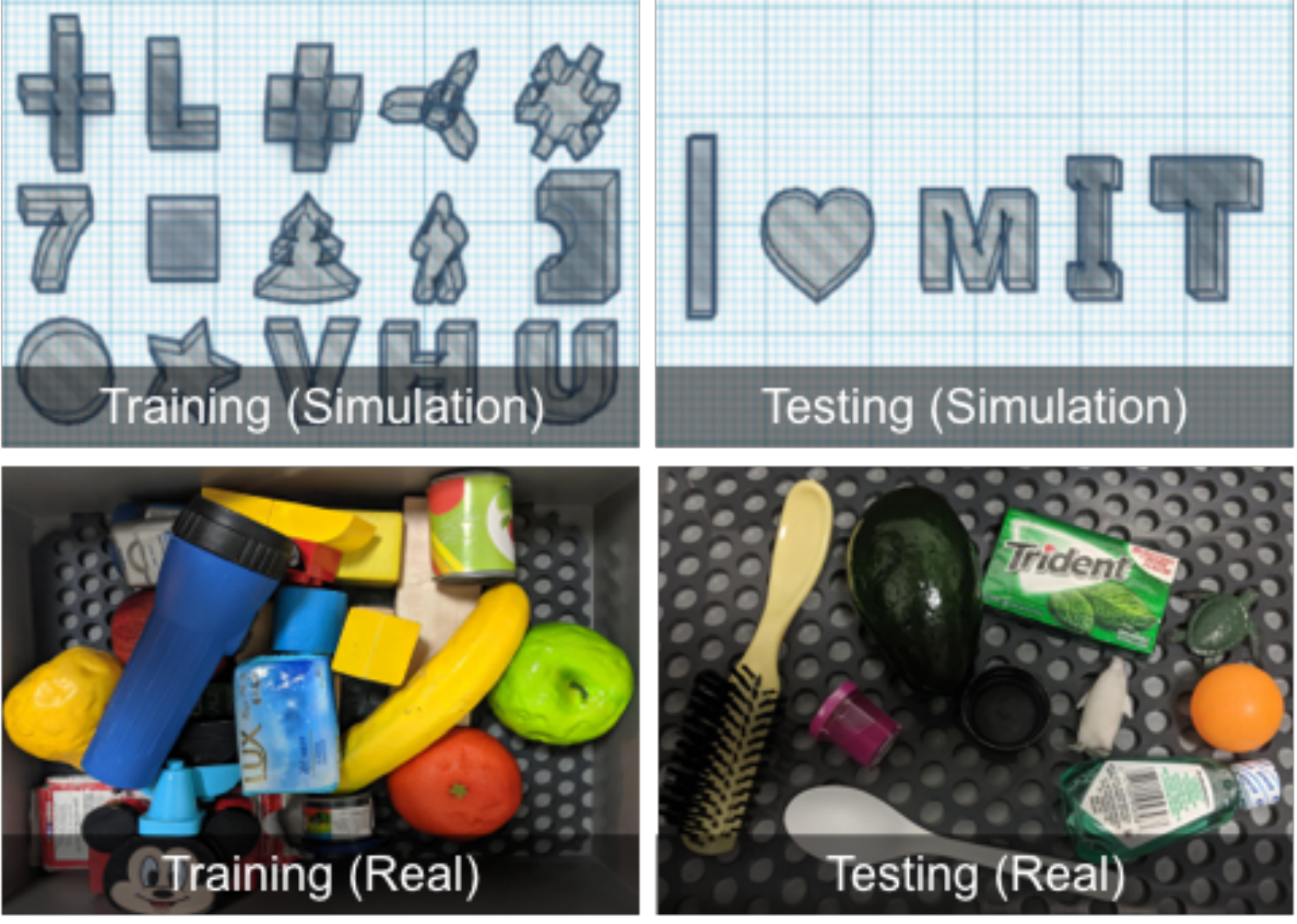}\vspace{-2mm}
    \caption{ \textbf{Objects used in experiments.} We show objects in simulated (top) and real (bottom) experiments, split by training objects (left) and testing objects (right) which are unseen during training. }
    \label{fig:objects}
    \vspace{-4mm}
\end{figure}

\begin{figure}[t]
    \centering
    \includegraphics[width=\linewidth]{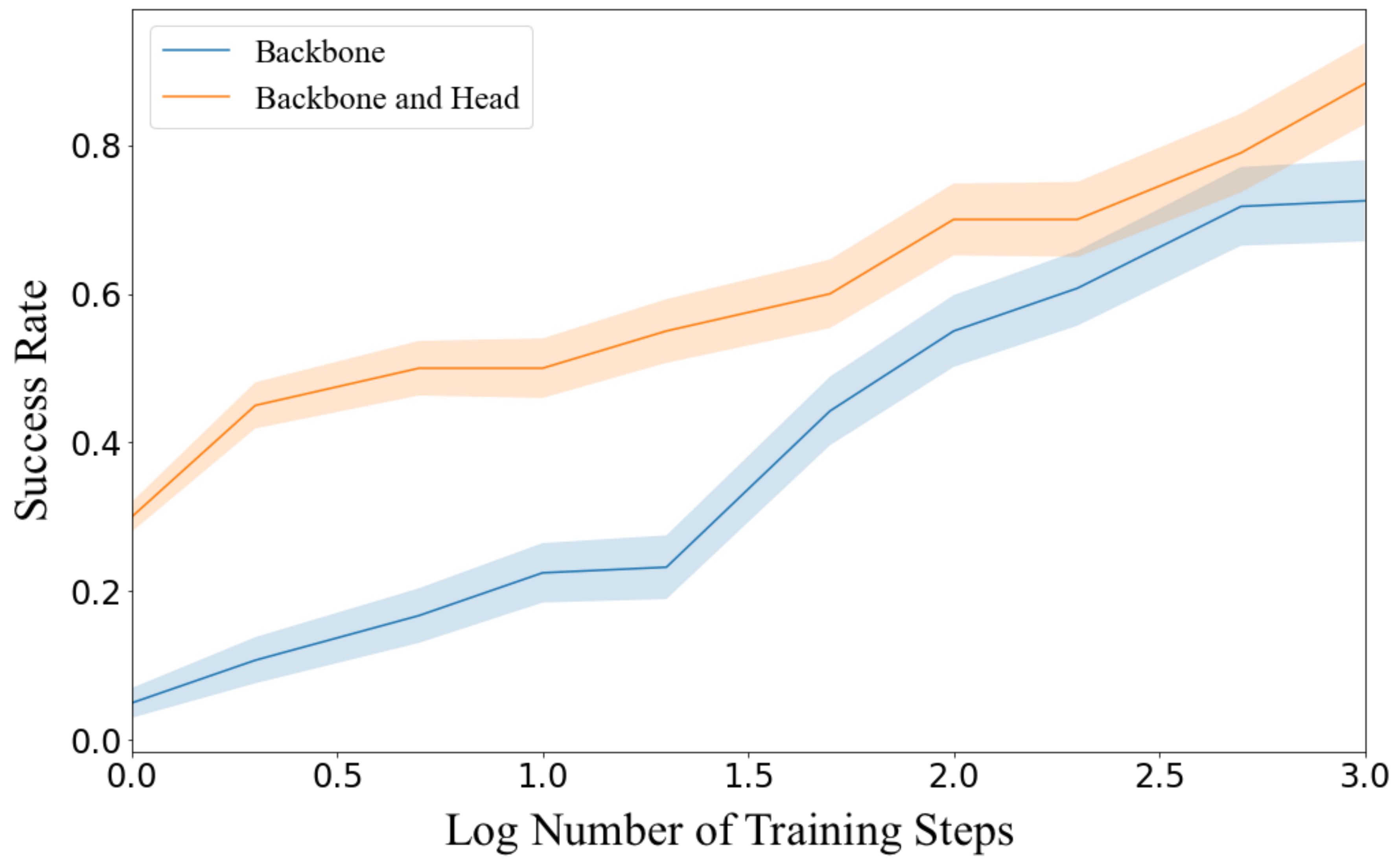} \vspace{-5mm}
    \caption{{\bf Performance of transfer learning.} Transferring entire model (backbone + head) achieved a success rate of 86\% while partial model (backbone only) merely achieved 71\%. We report the mean and 95\% confidence interval of the performance over five runs.
    %\caption{{\bf Simulation results of transferring partial (Backbone) and entire vision model (Backbone and Head) for affordance learning Tr.} 
    % \todo[inline]{if have time, make the x-ticks "1, 2, 5, 10, 20, ..." or "10^1, 10^2, ..."}
    }
    \label{fig:transfer_parts_grasping_sim}
    \vspace{-4mm}
\end{figure}

% We conducted a series of experiments in both simulated and real settings to study the problem of transferring trained vision models for learning grasping and suction. Here we start by introducing our

We executed a series of experiments in both simulated and real settings (conducted in isolation) to evaluate how well trained vision models can contribute to the performance of manipulation affordance models -- in particular, for suction and grasping. The goal of our experiments is three-fold: 1) to investigate whether transferring only a part of the network (\ie backbone) versus the entire network (\ie including both backbone and head) leads to improved manipulation performance, 2) to benchmark the value of various visual task(s) by how much they contribute to manipulation performance, and 3) to determine which large-scale real-world datasets result in the most improvement for manipulation performance.

\noindent \textbf{Evaluation protocol.} We evaluate the performance of our manipulation affordance models for suction and grasping by measuring picking success rate $\frac{\mathrm{\#~successful~picks}}{\mathrm{\#~picking~attempts}}$ \ie the rate which an object remains in the end-effector after executing the respective primitive.
% We adopt picking success rate, the rate which an object remains in the end-effector after executing the picking primitive, as our evaluation metric.
%
% To evaluate the progress objectively, we reset the environment to a pre-defined canonical setting before each evaluation.
%
% During evaluation, gradient is disabled and the model is thus fixed.
%
In all experiments, we report the total picking success rate per test run after executing a maximum of 25 picking attempts with seen objects used for training and 15 picking attempts with unseen objects. If the robot successfully picks up all available objects before reaching the maximum count of picking attempts, the test run ends. Prior to each test run, we reset the environment to a pre-defined canonical cluttered arrangement of objects, unseen during training. During evaluation, model gradients are disabled.

\subsection{Simulation Experiments} \label{sec:sim_experiments}

\begin{figure*}[ht!]
    \centering
    \includegraphics[width=\linewidth]{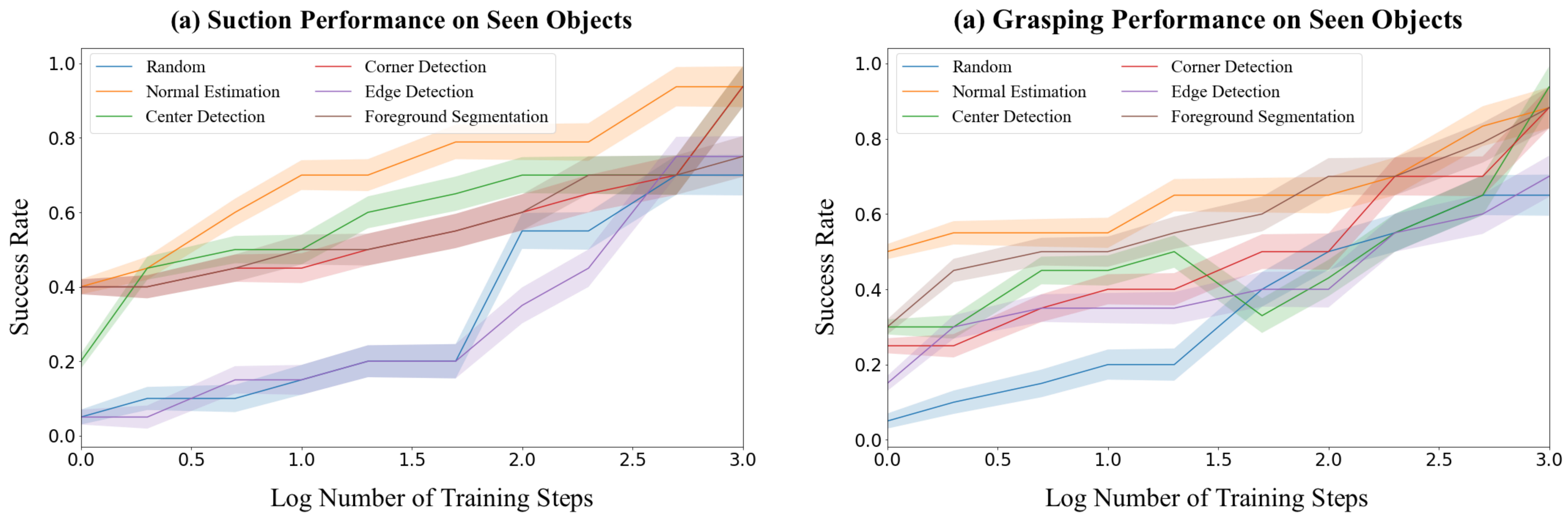}\vspace{-3mm}
    \caption{{\bf Simulation results of suction and grasping with models pre-trained on different vision tasks.}
    We report the mean and 95\% confidence interval of the performance over five runs.
        % \todo[inline]{if have time, make the x-ticks "1, 2, 5, 10, 20, ..." or "10^1, 10^2, ..."}
    }
    \label{fig:suction_grasping_sim}
    \vspace{-3mm}
\end{figure*}

% simulation experiments first: 1) backbone + head, 2) which visual tasks?
% realworld experiments second: 1) validate sim, 2) coco is amazing, 3) compare training from scratch
% analysis should be interleaved above

%
%\noindent \textbf{Setup.}

Our simulation environment in PyBullet uses a UR5 robot arm with a Robotiq 2F-85 gripper. We simulate picking experiments with 20 different 3D object models: 15 seen during training and 5 unseen for testing (illustrated in~\figref{objects}). These objects were chosen to maximize shape diversity. Two copies of each 5 object (10 in total) are randomly colored, scaled, and dropped into the bin during training and testing. For our simulation experiments, we follow Zeng et al.~\cite{zeng2019tossingbot} to use a 7-layer fully convolutional residual network as backbone network architecture.

%The simulation environment is built using PyBullet. We use 20 different objects: 15 seen during training and 5 unseen for testing as shown in the top row of~\figref{objects}. Training objects are chosen in order to increase shape diversity. Multiple copies of each object (10 in total) is randomly colored, scaled, and dropped into the bin during training and testing. In our simulation experiments, we adopted the backbone architecture from Zeng et al.~\cite{zeng2019tossingbot}.
%

%
%

\noindent\textbf{Transfer partial or entire vision model?}~\label{which_to_transfer}
Our first experiment compares transferring pre-trained weights from a partial backbone model versus pre-trained weights from both the backbone and vision head. This comparison is interesting because substantial prior work in computer vision uses the former pre-training paradigm for better performance -- we test to see if this assumption still holds true for vision-to-manipulation transfer. Specifically, we train a foreground segmentation model in simulation (for which ground truth is available), use its weights to initialize a grasping affordance model, then fine-tune the affordance modes through trial and error. \figref{transfer_parts_grasping_sim} plots the picking success rates over training steps during fine-tuning using each of the aforementioned pre-training paradigms.

Interestingly, we observe that initializing our manipulation affordance models using pre-trained weights from both the backbone model and vision head yields the strongest overall picking performance. This is likely because the pre-trained weights from the vision head help to positively bias grasp samples predicted by the manipulation model towards foreground objects. This decreases the number of randomly predicted grasps in the background (as would be the case if the head was randomly initialized) early in training, and thus improves exploration and sample efficiency.

%First, we investigate whether the common practice of transfer learning for vision tasks, \ie only transferring features, is ideal when target tasks receive supervisions from interaction.
%
%To answer the question, we train a foreground segmentation model in simulation and transfer it to learn suction and grasping. We show the training curves of model which only transfers the latent features and model which transfers both latent features and prediction head in Fig. X.
%
%We found that when transferring both features and prediction head, models can leverage the visual prediction to better explore the possible interactions instead of acting randomly at the early stage of training, which drastically improve the sample efficiency.

% \begin{table}[t]
% \begin{center}
% \begin{tabular}{ l|r|r } 
% \hline
% Vision Tasks & Suction & Grasping \\ \hline 
% Random    & 0.33 & 0.46 \\
% Foreground segm.    & - & 0.53 \\
% Normal estimation    & 0.46 & - \\
% ImageNet   & 0.40 & 0.46 \\
% COCO-backbone        & 0.77 & 0.83 \\
% COCO        & 0.91 & 0.90 \\
% \hline
% \end{tabular}
% \end{center}
% \caption{Real performance on unseen objects.}
% \label{tab:sim_test}
% % \vspace{-2mm}
% \end{table}

\noindent\textbf{Which vision tasks to transfer?}~\label{what_to_transfer} We next investigate the benefits of pre-training on various different vision tasks, for which we can obtain ground truth training labels in simulation. These tasks include: 1) object center detection, 2) edge detection, 3) corner detection, 4) foreground segmentation, and 5) flat surface detection. We transfer pre-trained weights from these tasks to suction and grasping affordance models, then report their picking success rates while fine-tuning with trial and error on the training objects in simulation in~\figref{suction_grasping_sim}. We also execute test runs with novel unseen objects using the fine-tuned affordance models, and report their performance in Table \ref{tab:sim_test}.

\begin{table}[h!]
\caption{Simulation performance on unseen objects.}
\vspace{-6mm}
\begin{center}
\begin{tabular}{ l|r|r } 
\toprule
Vision Tasks & Suction & Grasping \\ \midrule 
Random    & 0.71 & 0.71 \\
Normal estimation   & 1.00 & 0.91 \\
Center detection        & 0.83 & 0.91 \\
Corner detection        & 0.83 & 0.91 \\
Edge detection    & 0.90 & 0.83 \\
Foreground segm.    & 0.83 & 0.91 \\
\bottomrule 

\end{tabular}
\end{center}
\label{tab:sim_test}
\vspace{-4mm}
\end{table}

%
% Second, we benchmark the benefit of pre-training on different vision tasks for learning robotics tasks in simulation. We pre-train vision models on the following tasks: 1) center detection, 2) edge detection, 3) corner detection, 4) foreground segmentation, and 5) flat surface detection. Then, we transfer these models to learn suction and grasping.
%
%The training curve for seen objects are shown in~\figref{suction_train_sim} and~\figref{grasping_train_sim}. The generalization results to unseen objects are shown in table X.
%

From these results, we find that for the suction affordance model, pre-training on the vision task of flat surface detection drastically improves its sample efficiency. This is as expected since suction performs better (both in simulation and in real-world) on flat object surfaces than on other geometries. For the grasping affordance model, pre-training on foreground segmentation and flat surface detection outperforms other variants, while the former task enables the model to generalize better to unseen objects. This is surprising since we expected corner and edge detection to provide features that are more relevant for grasping. However, our results indicate that the grasping affordance model benefits most from understanding the distinction between foreground and background pixels.

%
%We found that for suction, pre-training on flat surface detection dramatically improves the sample efficiency. For grasping, pre-training the model on foreground segmentation and flat surface detection outperform other variants, while the former generalizes better to unseen objects better.
%

\subsection{Real-world Experiments}

Our real-world setup consists of a UR5 robot arm, equipped with an RG2 gripper for grasping, a 3D-printed end effector with a B-Bellows DURAFLEX suction cup for suction, and a statically mounted Intel RealSense D415 overlooking a bin of objects from the side. The camera captures $1280 \times 720$ RGB-D images and is localized with respect to the robot base using an automatic calibration procedure from \cite{zeng2018learning}. Our collection of objects includes 25 different toy blocks, fake fruit, decorative items, and office objects (see bottom left of~\figref{objects}) during training. We use 10 additional objects (see bottom right of~\figref{objects}) for evaluating our model on its generalization to unseen objects. For our real-world experiments, we use the region proposal network (RPN) model architecture in~\cite{detectron2018}. We adopt ResNet-50 FPN~\cite{lin2017fpn} as the backbone model and prediction layers on $P2$ feature level as the head model. The multi-anchor predictions are collapsed by max-pooling on confidence scores to generate a single heatmap. To investigate the effects of vision tasks, we also include RPNs that are pre-trained on foreground segmentation and normal estimation as baselines. The data consists of 1000 cluttered observations with labels computed from depth images.

% During training, we use an UR5 arm to pick a collection of 25 different toy blocks, fake fruit, decorative items, and office objects (see bottom left of~\figref{objects}) during training. To evaluate whether our model can generalize to unseen objects, we select 10 more objects for testing  (see bottom right of~\figref{objects}). UR5 arm's end effector is an RG2 gripper for grasping and a 3D-printed case equipped with a B-Bellows DURAFLEX suction cup for suction. For perception data, we capture $1280 \times 720$ RGB-D images using a calibrated Intel RealSense D415 statically mounted overlooking the bin of objects from the side. The camera is localized with respect to the robot base using an automatic calibration procedure from [xxx]. In our real world experiments, we use region proposal network (RPN) architecture in~\cite{detectron2018}. We adopt ResNet-50 FPN~\cite{lin2017fpn} as the backbone and score predictions on $P2$ feature level as head. The multi-anchor predictions are collapsed by max pooling on confidence scores to output single heatmap.

\begin{figure*}[t!]
    \centering
    \includegraphics[width=\linewidth]{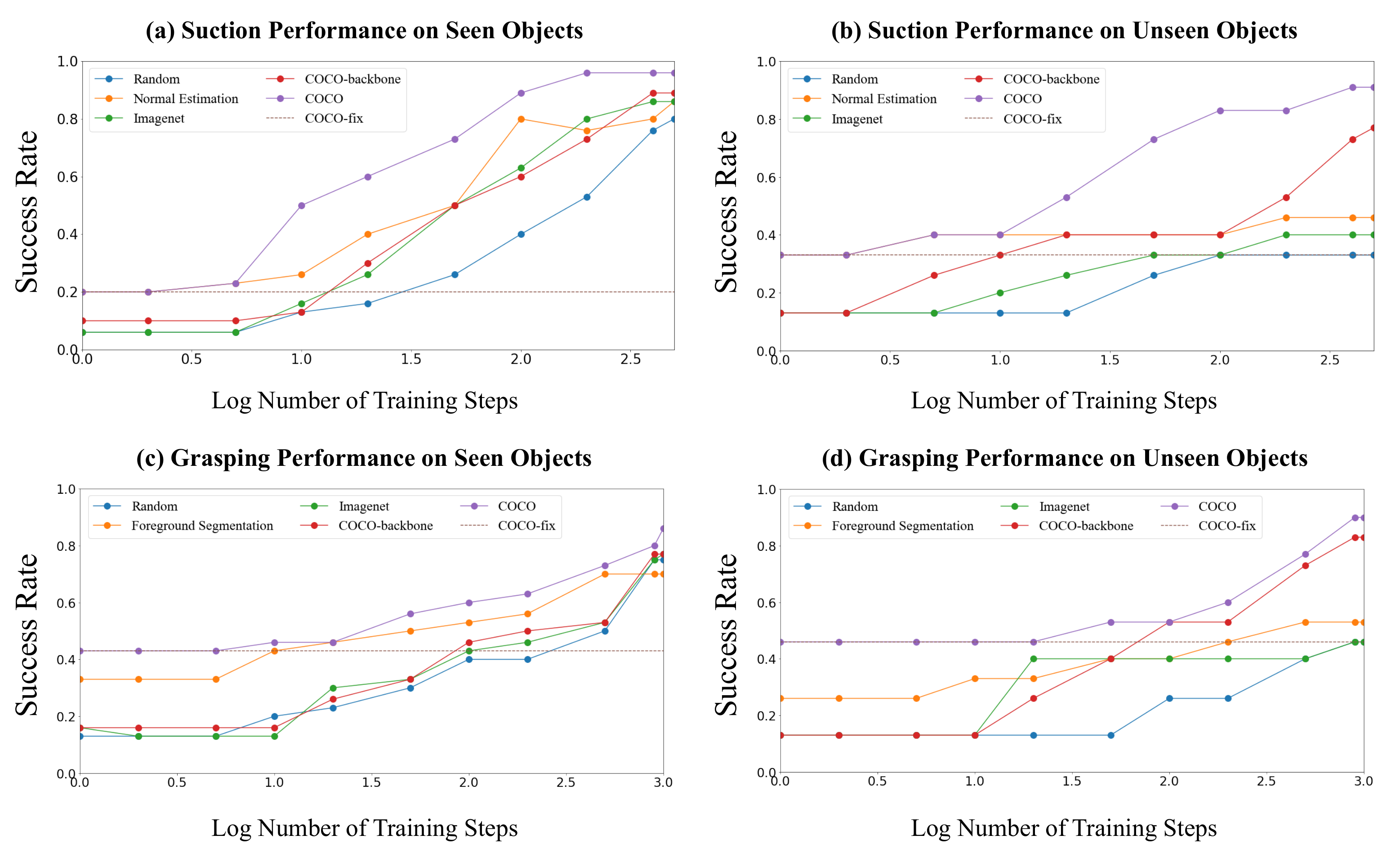}\vspace{-5mm}
    \caption{{\bf Real-world results of suction and grasping with different pre-trained models.}
    }
    \vspace{-5mm}
    \label{fig:suction_real}
\end{figure*}

%
\noindent\textbf{Can off-the-shelf vision models help?} In our real-world experiments, we train vision models with the best-performing vision tasks from our simulation experiments (Section \ref{sec:sim_experiments}) and compare them with random initialization to verify our findings in simulation. 
% In our real-world experiments, we train the vision models of the best-performing vision tasks in simulation and compare it with random initialization to verify our findings from~\ref{which_to_transfer} and~\ref{what_to_transfer}. 
%
% We then explore whether off-the-shelf vision models can serve as good initialization for learning suction and grasping affordances. 
%
We then explore whether off-the-shelf vision models can serve as good initialization for learning suction and grasping affordances. These off-the-shelf pre-trained models include: a) ImageNet, b) COCO-backbone, c) COCO, and d) COCO-fix, where both ImageNet and COCO-backbone only transfer backbone network weights. COCO and COCO-fix transfer both backbone and head network weights but COCO-fix is not trained.
% To transfer the off-the-shelf vision models, we benchmark the following pre-trained features: a) ImageNet, b) COCO-backbone, and c) COCO, where both ImageNet and COCO-backbone only transfers the latent features.
%
Note that off-the-shelf vision models are not fine-tuned on the target domain and thus don't require additional labels.
We transfer pre-trained weights from these tasks to learn suction and grasping affordance models, then report picking success rates while fine-tuning with trial and error on the training objects in~\figref{suction_real}. We also execute test runs with novel unseen objects (including transparent objects \cite{sajjan2019cleargrasp}) using the affordance models and report their performance in ~\figref{suction_real}.
These results highlight three key observations: 1) we validate that our findings in simulation apply to our real-world setting,
2) off-the-shelf vision models generalize better than vision models trained in the target environment (COCO-fix vs. Random on unseen objects, ~\figref{suction_real}b,d)
3) models initialized with COCO features perform better than models initialized with ImageNet features.
Finding 2) is encouraging as it shows that learning from diverse offline observations (\ie COCO dataset) can improve both sample efficiency and generalization when learning manipulation affordances.
%
%For finding 3), one possible explanation is that features learned from object detection focus more on shape, and thus perform better compared to other model that focuses more on texture (e.g., ImageNet).
For finding 3), a possible explanation is that the features learned from object detection dataset are able to localize objects, which is crucial for manipulation; whereas models trained with image classification lose this spatial information.
%

%
% \begin{figure*}[]
%     \centering
%     \includegraphics[width=\linewidth]{figures/grasping_real.pdf}
%     \caption{{\bf Real-world results of grasping.} 
%     %
%     }
%     \label{fig:grasping_real}
% \end{figure*}
%

\begin{figure}[t]
    \centering
    \includegraphics[width=\linewidth]{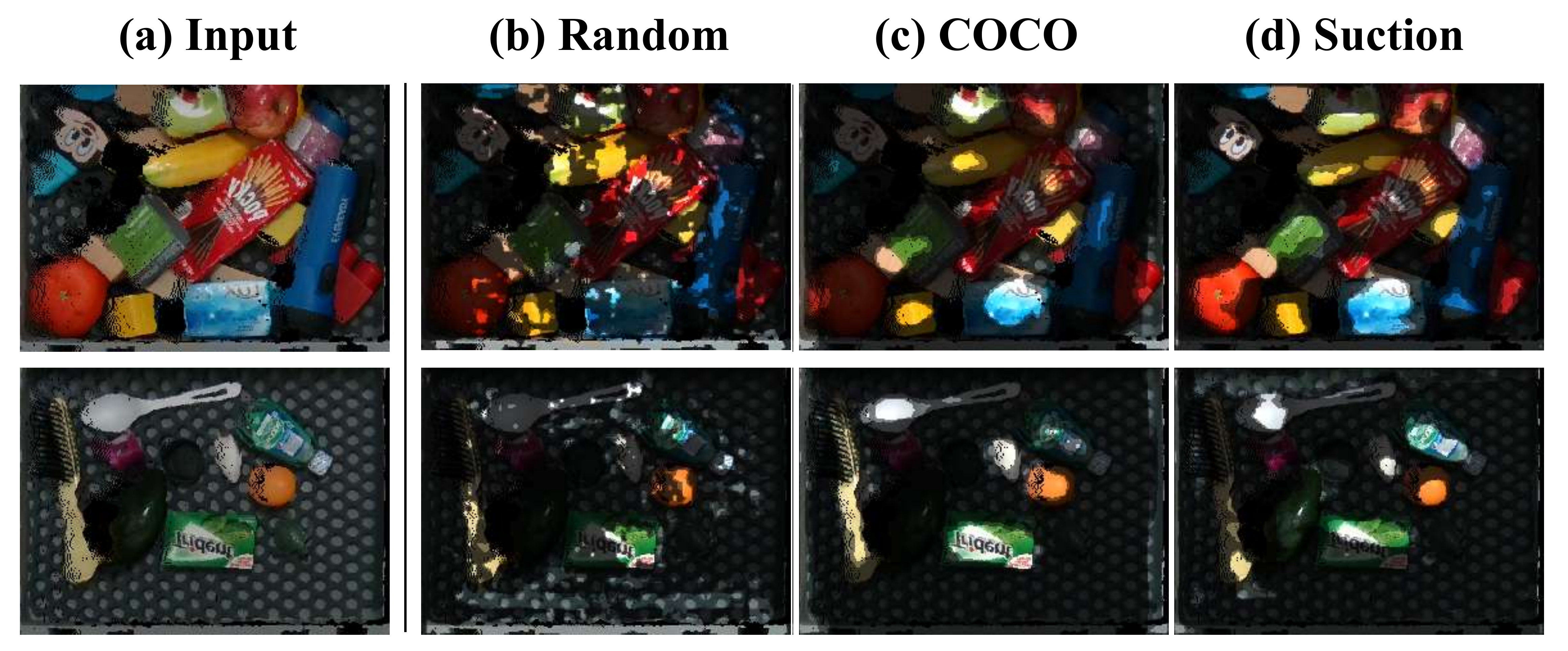}\vspace{-3mm}
    \caption{{\bf Affordances predicted by different models.}
    We show the affordances of suction predicted by different models given input from (a). (b) Random refers to a randomly initialized model. (c) COCO is the modified RPN model. (d) Suction is a converged model learned from robot-environment interactions.
    }
    \vspace{-6mm}
    \label{fig:visualization}
\end{figure}

\noindent\textbf{Visualization analysis.}
To understand why vision tasks transfer well for learning affordances, we show the heatmap visualization of vision and affordance models (where hotter regions indicate higher predicted values).
As shown in~\figref{visualization}, the Random and ImageNet models output random heatmaps since the prediction head is randomly initialized. The heatmap of Normal and COCO models show strong object-centric activations, with an output distribution very similar to that of trained suction models. The heatmap visualization supports our intuition that vision and affordance tasks share strong similarities.

% a transferred model which encodes objcet-centric prior naturally focuses on the object from the very start of training while model with random weights spread uniform focus in the scene.
%As shown in~\figref{visualization}, a transferred model which encodes objcet-centric prior naturally focuses on the object from the very start of training while model with random weights spread uniform focus in the scene.
\section{CONCLUSION AND FUTURE WORK}
In this work, we explore the extent to which pre-trained vision models (from passive observations) can benefit the process of learning affordance models (for active manipulation).
We present a method that can significantly reduce the number of interactions required for learning manipulation policy by first learning visual predictions. A wide variety of visual tasks is studied in this work and we show that learning visual predictions not only provide better features but also provide good initial manipulation strategy. The real-world experiments show that off-the-shelf vision model provides generic vision features across domains, improves both training speed and final performance for learning manipulation in a new environment. We imagine the benefit of visual pre-training is not only limited for affordance based manipulation model. A future research direction is to apply vision-guided exploration to action model training with reinforcement learning, such as~\cite{kalashnikov2018qtopt}, where the data efficiency is the bottleneck of learning.
\section*{Acknowledgements}
We would like to thank the team members of Robotics at Google for insightful discussions and technical support. 

\clearpage
\bibliographystyle{IEEEtran}
\bibliography{IEEEexample}

\end{document}